\PassOptionsToPackage{unicode}{hyperref}
\PassOptionsToPackage{hyphens}{url}
\documentclass[3pt, twocolumn]{article}
\usepackage{xcolor}
\usepackage{amsmath,amssymb}
\usepackage{graphicx}
\usepackage{float}
\usepackage{comment}
\usepackage{booktabs} 
\usepackage{tabularx}
\usepackage{ragged2e}
\usepackage[a4paper, left=2cm, right=2cm, top=2cm, bottom=2cm]{geometry}
\usepackage{enumitem}
\usepackage{iftex}
\usepackage{authblk}
\ifPDFTeX
  \usepackage[T1]{fontenc}
  \usepackage[utf8]{inputenc}
  \usepackage{textcomp} 
\else 
  \usepackage{unicode-math} 
  \defaultfontfeatures{Scale=MatchLowercase}
  \defaultfontfeatures[\rmfamily]{Ligatures=TeX,Scale=1}
\fi
\usepackage{lmodern}
\ifPDFTeX\else
\fi
\IfFileExists{upquote.sty}{\usepackage{upquote}}{}
\IfFileExists{microtype.sty}{
  \usepackage[]{microtype}
  \UseMicrotypeSet[protrusion]{basicmath} 
}{}
\makeatletter
\@ifundefined{KOMAClassName}{
  \IfFileExists{parskip.sty}{%
    \usepackage{parskip}
  }{
    \setlength{\parindent}{0pt}
    \setlength{\parskip}{6pt plus 2pt minus 1pt}}
}{
  \KOMAoptions{parskip=half}}
\makeatother
\usepackage{color}
\usepackage{fancyvrb}

\DefineVerbatimEnvironment{Highlighting}{Verbatim}{commandchars=\\\{\}}

\usepackage{longtable,booktabs,array}
\usepackage{calc} 
\usepackage{etoolbox}
\makeatletter
\patchcmd\longtable{\par}{\if@noskipsec\mbox{}\fi\par}{}{}
\makeatother
\IfFileExists{footnotehyper.sty}{\usepackage{footnotehyper}}{\usepackage{footnote}}
\makesavenoteenv{longtable}
\providecommand{\tightlist}{%
  \setlength{\itemsep}{0pt}\setlength{\parskip}{0pt}}
\usepackage{bookmark}
\IfFileExists{xurl.sty}{\usepackage{xurl}}{} 
\hypersetup{
  hidelinks,
  pdfcreator={LaTeX via pandoc}}

\title{Attention In Geometry: An Adaptive Density Field (ADF) with
FAISS-Accelerated Spatial
Attention}
\author[1]{Zhaowen Fan}
\author[1]{Yunxiang Han\thanks{Corresponding author: hannuaa@126.com}}

\affil[1]{School of Computer, SICHUAN UNIVERSITY, CHENGDU, CHINA, 610065}

\date{}

\begin{document}
\maketitle

\begin{abstract}
Spatial computation in geographic systems increasingly requires
query-conditioned, local, interpretable aggregation under metric constraints.
Many classical approaches rely on global
summation and treat approximation as an implementation concern,
limiting interpretability and scalability in large-scale. We propose
Adaptive Density Field (ADF), a geometric
attention framework that formulates spatial aggregation as a
query-conditioned, metric-induced attention operator in continuous space.
Given a set of labelled spatial points with associated scalar scores, ADF
defines a continuous intensity field over space. For a given query
location, the field value is obtained via a local adaptive Gaussian kernel
mixture centered on the query's nearest neighbors, where kernel bandwidths
are modulated by point-specific scores to evaluate local aggregated influence.
Additionally, approximate nearest-neighbor search is introduced, enabling scalable
execution while preserving locality. The proposed ADF bridges concepts from
adaptive kernel methods, classical GIS methods, and the attention mechanisms
by reinterpreting spatial influence as geometry-embedded attention, grounded
in physical distance rather than learnt latent projections.
The proposed framework is formulation-level rather than algorithm-specific,
allowing flexible kernel choices, score-to-bandwidth mappings and
approximation parameters. This approach provides a unifying
perspective on spatial influence modeling that emphasizes structure,
scalability, and geometric interpretability, with relevance to geographic
information systems and spatial machine learning.
\end{abstract}

\section*{Keywords}
Adaptive Density Fields; Kernel Density Estimation; Spatial Point Patterns;
Approximate Nearest Neighbors; FAISS

\section{Introduction}
\label{introduction}

Geographic Information Science (GIScience) is a multidisciplinary field
\cite{goodchild1992geographical} concerned with developing analytical and
predictive methods for specific tasks involved with geographic data. While
recent advances in computer science have been dominated by the thriving of
large language models \cite{bommasani2021opportunities}, people often ignore
that a large-scale spatial systems also require a query-conditioned,
interpretable aggregation method, like the ones proposed for large language
models, that respects metric locality while remaining computationally
feasible. Such requirements arise naturally in a range of GIScience
scenarios, including on-demand accessibility estimation in urban
environments, real-time assessment of spatial influence around moving
objects (e.g., vehicles, drones, or trajectories), point-of-interest impact
modeling, adaptive exposure analysis in environmental monitoring, and
interactive spatial querying in large-scale geographic databases
\cite{tian2025research, luo2023spatial, miller2015data,
shekhar2007encyclopedia}. In these settings, spatial influence must be
evaluated locally at arbitrary query locations, often under strict latency
and memory constraints, which makes global or batch-based aggregation
methods impractical \cite{shekhar2012benchmarking}.

Classical adaptive KDE \cite{silverman1986density} and k-NN density
estimator \cite{cover1967nearest} provide a statistical foundation for
modern variable bandwidth smoothing, but they are typically global or
batch estimators and do not treat approximation as part of the operator
definition \cite{biau2011weighted}. Currently, on one hand several recent
applied pipelines have attempted to combine ANN acceleration with
KDE for large datasets for better scalability \cite{thompson2022ancient,
koylu2019deep, zhang2017gpu}, and on the other hand, query-driven KDE
variants have also been proposed for influence region computation and other
domains \cite{alghushairy2020genetic, govorov2025exploration}. However,
these approaches either assume full density estimation or lack explicit
score-modulated attention semantics and intrinsic sparsification, which
limits their ability to explicitly encode query-conditioned sparsification,
score-dependent influence modulation, and operator-level interpretability.
In particular, approximation is typically treated as an external
optimization rather than an intrinsic component of the aggregation operator,
and kernel influence is not formulated as an attention-like mechanism that
directly reflects spatial locality and point-specific importance. As a
result, while being robust and well-established, existing methods
unavoidably offer limited flexibility in trading off locality,
interpretability, and computational efficiency within a unified formulation.

In this paper, we propose the Adaptive Density Field (ADF), a
geometry‑embedded attention operator that makes locality, sparsification,
and approximation first‑class design choices. In particular, we consider
the problem of constructing a continuous influence field  \(F(x)\) that
reflects the accumulated effect of nearby points of interest (POIs) under
physical distance constraints. This framework formulates spatial aggregation
as a query-conditioned, metric-driven weighted sum. For any query location
\(x \in \mathbb{R}^3\) (Earth-Centered, Earth-Fixed coordinates), ADF selects
a local neighbor set, assigns each neighbor a score-modulated kernel, and
aggregates the contributions to produce \(F(x)\), using the FAISS inverted
file indices \cite{johnson2017faiss} for acceleration.  This perspective is
related to kernelized and geometry-aware attention mechanisms, but differs
in that attention weights are induced directly by physical distance rather
than learned projections \cite{bronstein2021geometric,
choromanski2020rethinking}.

This work is structured as follows: In Section~\ref{framework}, we introduce
the basic methodology and formal definition of the Adaptive Density Field
(ADF), positioning the framework as a conceptual contribution to GIScience.
In Section~\ref{flight-trajectory-poi-extraction}, we present a practical
instantiation that combines ADF with ANN indices for POI detection.
In Section~\ref{ablation-study} we evaluate the robustness and
efficiency of the proposed framework, and demonstrate the selection of
parameters. In Section~\ref{comparative-evaluation-and-related-works} we
discuss the relationship and differences between the proposed method and
several classic and well-known GIS method. In Section~\ref{discussion},
we discuss the implications of ADF as a unifying operator-level perspective
on spatial influence modeling, emphasizing geometric grounding, explicit
approximation, and scalability. Finally, Section~\ref{conclusion} summarizes
the main conclusions and outlines future work.

\section{Framework}
\label{framework}

\subsection{Overview}
\label{overview}

The Adaptive Density Field (ADF) is a geometric attention operator designed
to unify locality, sparsity, and approximation in continuous spatial
aggregation. Rather than focusing on implementation or dataset-specific
optimizations, the framework formalizes spatial influence as an
operator-level construct: given a query location \(x \in \mathbb{R}^3\)
(ECEF coordinates), ADF aggregates contributions from nearby points of
interests (POIs) through score-modulated kernels and a metric-induced
weighting scheme per query.

This section provides a conceptual roadmap of the framework. We first
define the operator mathematically, discuss the score-to-bandwidth
mappings, and approximation mechanisms, and then present the usage of
FAISS for efficient neighbor retrieval as an approximate nearest-neighbor
(ANN) method. Finally, we evaluate the time and memory complexity of the
proposed method, highlighting the flexibility of the approach and its
suitability for a range of GIScience applications.

\subsection{Data Preparation}
\label{data-preparation}

\subsubsection*{Suggested data structures: positions and scores}
\label{suggested-data-structures-positions-and-scores}

In the experiments, POIs are derived from a physics-informed trajectory
analysis pipeline (detailed in the Appendix~\ref{appendix}); each POI is associated
with a spatial location and a scalar score reflecting deviation magnitude.
The details of this pipeline are not central to the ADF formulation and are
therefore omitted. ADF operates on arbitrary spatial point sets with
associated scores, defined as follows:

\begin{figure}[htbp]
    \centering
    \includegraphics[width=\columnwidth]{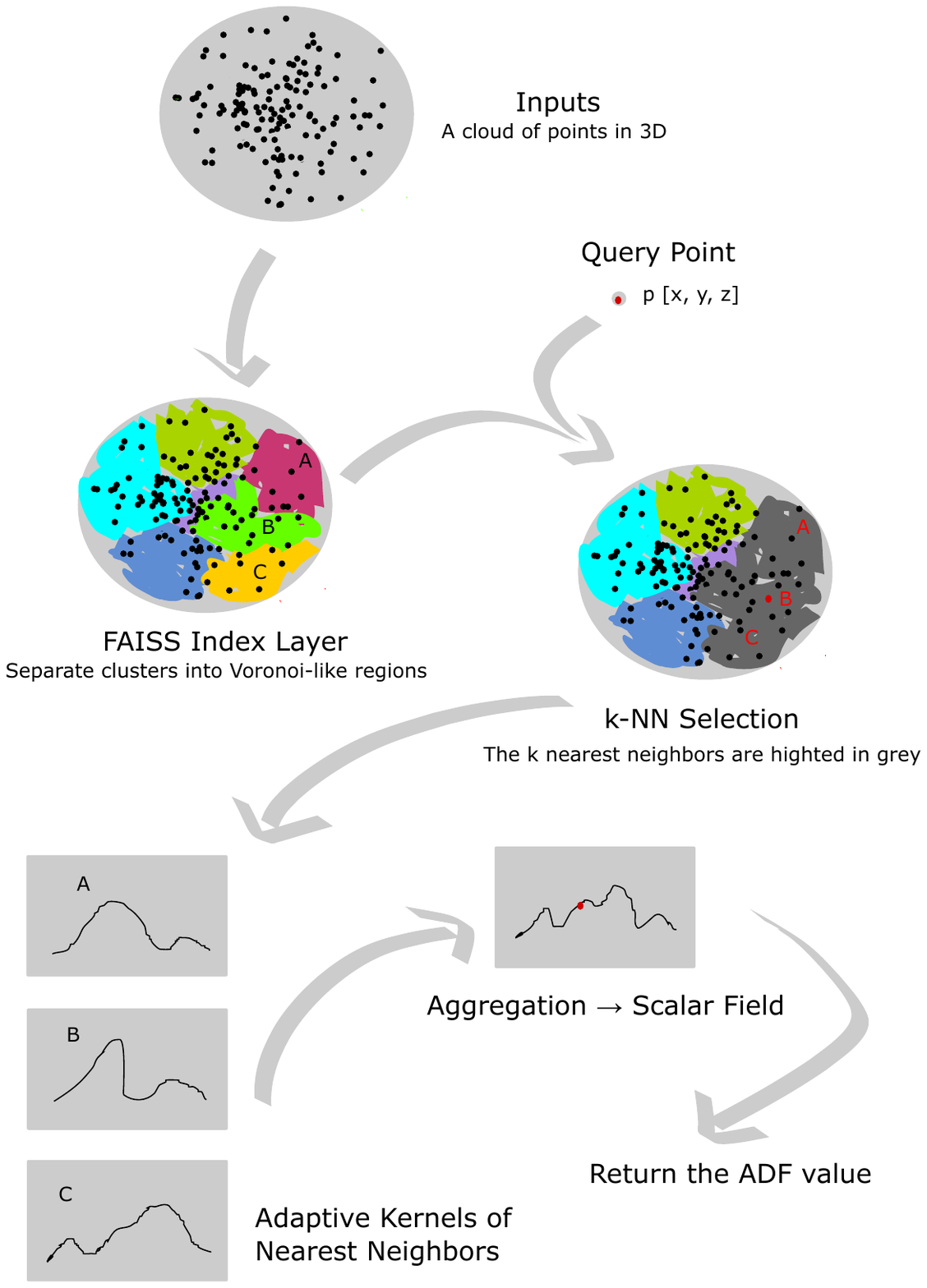}
    \caption{An overview of the proposed framework}
    \label{fig:Fig.1}
\end{figure}

(i) \(\mathbf{x}_i \in \mathbb{R}^3\) is the ECEF position of POI \(i\)

(ii) \(s_i\) is the score (weight) of POI \(i\). Formally, the input is
defined as \(\{\mathbf{x}_i, s_i\}_{i=1}^n\), where
\(\mathbf{x}_i \in \mathbb{R}^3\) and \(s_i \in \mathbb{R}\). The
score \(s_i\) represents an application-dependent measure of influence,
saliency, or importance; its semantic interpretation is not constrained
by the ADF formulation.

\subsection{Methodology}
\label{methodology}

Given the POI data, we construct the ADF as follows:

\subsubsection{FAISS IVF Index: Accelerating Neighbor Search}
\label{faiss-ivf-index-accelerating-neighbor-search}

For the proposed framework, FAISS is used purely as a computational
accelerator to retrieve local neighborhoods efficiently from large
POI sets. Parameters such as k and nprobe control access to sufficient
local context rather than defining the ADF model itself (we will discuss
the parameter selection in Section~\ref{ablation-study}). In practice,
results were stable once neighborhoods exceeded a minimal size, and all
experiments used a fixed, conservative configuration. For each POI data
imported, we assign the IVF index \cite{jegou2011product}, and the
procedure for high-dimensional indexing is formalized as follows:

\paragraph{Training}

(i) FAISS runs k-means with \(n_{\text{list}} = 4096\)
clusters in \(\mathbb{R}^3\).

(ii) It learns centroids
\(\{\mathbf{c}_j\}_{j=1}^{4096}\).

This formalizes the cluster assignment, though standard FAISS documentation
may be consulted for full details.

\paragraph{Assignment}

Each vector \(\mathbf{x}_i\) is assigned to its nearest centroid:
\begin{equation}
\ell(i) = \arg\min_{1 \le j \le 4096} \|\mathbf{x}_i - \mathbf{c}_j\|^2.
\end{equation}
and it is stored in inverted list \(L_{\ell(i)}\).

\paragraph{Search}

For a query \(\mathbf{x}\), FAISS finds the 16 nearest centroids
\(j_1, \dots, j_{16} \): indices of the 16 closest
\(\mathbf{c}_j\) to \(\mathbf{x}.\) Then searches only in the union
of these lists: \(L_{j_1} \cup \dots \cup L_{j_{16}}\).

\subsubsection{ADF Model: An Adaptive Gaussian Mixture}
\label{adf-model-an-adaptive-gaussian-mixture}

\paragraph{Neighbor search}
\label{neighbor-search}

For a query point \(\mathbf{x} \in \mathbb{R}^3\) FAISS returns indices:
\(\{i_1, \dots, i_k\} = \text{k-NN}_{\text{IVF}}(\mathbf{x})\) in
approximate nearest‑neighbor sense. Then set: (i) Neighbor positions:
\(\mathbf{x}_{i_j} \in \mathbb{R}^3\); (ii) Neighbor scores: \(s_{i_j}\)

\paragraph{Differences}
\label{differences}

Mathematically, for each neighbor \(j = 1 \dots k\), we have:
\begin{equation}
\mathbf{d}_j = \mathbf{x}_{i_j} - \mathbf{x} \in \mathbb{R}^3.
\end{equation}

\paragraph{Adaptive bandwidth}
\label{adaptive-bandwidth}

Conceptually, higher score points are supposed to represent stronger
local influence, hence narrower kernels; lower scores should spread
influence more diffusely. Therefore, for each neighbor \(j\):
\begin{equation}
\label{eq:bandwidth}
\sigma_j = \frac{\sigma_0}{s_{i_j} + 10^{-6}}.
\end{equation}
\emph{The \(\epsilon = 10^{-6}\) term ensures numerical stability and avoids
singular bandwidths.} Therefore:

\begin{itemize}
\item
  High score \(s_{i_j} \implies\) smaller \(\sigma_j \implies\) more
  peaked kernel.
\item
  Low score \(s_{i_j} \implies\) larger \(\sigma_j \implies\) broader
  kernel. This is why it's \emph{adaptive}: kernel width depends on the
  score.
\end{itemize}

Then for each \(j\) we have:
\begin{equation}
\text{inv\_var}_j = \frac{1}{\sigma_j^2}.
\end{equation}

\begin{quote}
NOTE: The specific functional form relating scores to bandwidths is a
\emph{design choice}, not a defining property of ADF. ADF is defined by
(i) query-conditioned neighbor selection; (ii) metric-induced kernel
weighting; and (iii) scalable approximate execution. While this study employs
a specific reciprocal mapping, any monotonic or learned parameterization may
be substituted without altering the underlying framework.
\end{quote}

\paragraph{Quadratic form (Mahalanobis distance squared
\cite{mahalanobis1936distance})}
\label{quadratic-form-mahalanobis-distance-squared}

From Equation~\ref{eq:bandwidth}  we can derive the componentwise square
\(\mathbf{d}_j^2\).
Additionally, we have an array with shape \((k, 1)\),
broadcasts \(\text{inv\_var}_j\) across the 3 dimensions. So for each
neighbor \(j\):
\begin{equation}
\text{quad}_j
= \sum_{d=1}^3 (x_{i_j,d} - x_d)^2 \cdot \frac{1}{\sigma_j^2}
= \frac{\|\mathbf{x}_{i_j} - \mathbf{x}\|^2}{\sigma_j^2}.
\end{equation}
This is a derived Mahalanobis distance squared for an isotropic Gaussian
with variance \(\sigma_j^2\):
\begin{equation}
\text{quad}_j = (\mathbf{x} - \mathbf{x}_{i_j})^T \Sigma_j^{-1}
(\mathbf{x} - \mathbf{x}_{i_j}),
\end{equation}
where
\[\Sigma_j = \sigma_j^2 I_3, \quad \Sigma_j^{-1} = \frac{1}{\sigma_j^2}
I_3.\]

\paragraph{Gaussian kernel and ADF value}
\label{gaussian-kernel-and-adf-value}

For each neighbor \(j\), the kernel contribution is:
\begin{equation}
K_j(\mathbf{x}) = \exp\left(-\frac{1}{2} \text{quad}_j\right)
= \exp\left(-\frac{1}{2} \frac{\|\mathbf{x} - \mathbf{x}_{i_j}\|^2}
{\sigma_j^2}\right).
\end{equation}
In this work, we assume isotropic kernels with
\(\Sigma_j = \sigma_j^2 I_3\). While the ADF framework natively supports
anisotropic covariance matrices to account for directional trajectory
influence, for simplification, we applied isotropic kernels in this
study to establish a controlled baseline for evaluating the scalability and
partitioning stability of the ADF. This isotropic simplification isolates
the effects of score-modulated adaptation and scalable neighbor retrieval,
establishing a robust baseline before introducing directional anisotropy in
future trajectory-specific extensions. Then the \emph{ADF value} at
\(\mathbf{x}\) is assumed:
\begin{equation}
\label{eq:fx}
F(\mathbf{x}) 
= \sum_{j=1}^k s_{i_j} \, K_j(\mathbf{x})
= \sum_{j=1}^k s_{i_j} \exp\left(-\frac{1}{2} \frac{\|\mathbf{x} -
\mathbf{x}_{i_j}\|^2}{\sigma_j^2}\right).
\end{equation}
\begin{quote}
Note that \(F(x)\) is not normalized to integrate to unity and is
therefore not a probability density.
\end{quote}
Therefore, we claim that ADF is a \emph{finite Gaussian mixture}
\cite{bishop2006pattern}. Here, the term ‘Gaussian mixture’ refers to a
computational representation rather than a probabilistic generative model,
centered at the nearest neighbors, with:
\begin{itemize}
\item
  centers \(\mathbf{x}_{i_j}\);
\item
  weights \(s_{i_j}\);
\item
  bandwidths \(\sigma_j = \sigma_0 / (s_{i_j}+10^{-6})\).
\end{itemize}
We assume (i) \(s_i \ge 0\), or \(s_i\) is transformed via \(s_i
\gets \text{softplus}(s_i)\), (ii) \(\sigma_0\) or the score-to-bandwidth
mapping can be learned via gradient descent, (iii) \(\sigma_0\) has units
of meters and controls the physical spatial influence scale. This can be
interpreted as an adaptive kernel influence / intensity field in 3D ECEF
space.

The contribution of this work lies in the formulation of ADF as a
geometric attention mechanism with scalable execution. Kernel choices,
bandwidth parameterizations, and score transformations are intentionally
left flexible to emphasize the generality of the framework. In future
extensions tailored to directional flow data (e.g., aircraft trajectories),
anisotropic covariances can be induced by aligning principal axes with local
velocity vectors, fully realizing vector-aware geometric attention absent
in traditional VBKDE.

\subsubsection{Evaluating ADF at all POIs}
\label{evaluating-adf-at-all-pois}

From Equation~\ref{eq:fx}, for each POI position \(\mathbf{x}_i\):
\begin{equation}
\text{ADF}_i = \sum_{j=1}^k s_{i_j} \exp\left(-\frac{1}{2}
\frac{\|\mathbf{x}_i - \mathbf{x}_{i_j}\|^2}{\sigma_j^2}\right),
\end{equation}
where \(F(x)\) evaluated at POI \(x_i\) is denoted as \(ADF_i\), and
\(\{i_1,\dots,i_k\}\) are the \(k\)-nearest neighbors of \(\mathbf{x}_i\)
according to the IVF‑accelerated FAISS search. So the final result is a
\emph{scalar ADF field} defined on all your POIs:
\[\mathbf{x}_i \mapsto \text{ADF}_i.\]

\subsubsection{Approximation Error Acknowledgement}
\label{approximation-error-acknowledgement}

Since FAISS IVF performs approximate nearest-neighbor search, ADF(x) is
an approximation of the full kernel sum. However, locality of the
Gaussian kernel ensures distant errors contribute negligibly. In
practice, approximation quality is controlled by \(n_{\text{probe}}\)
and \(k\). Detailed numerical results for precision and recall are provided
in Sections~\ref{ablation-study} and
\ref{comparative-evaluation-and-related-works}.

\subsection{Computational Complexity Analysis}
\label{computational-complexity-analysis}

To assess the computational feasibility of the ADF framework for large-scale
trajectory analysis, we provide a theoretical analysis of its time and
memory complexity. Detailed benchmark experiments and comparison
are presented in Section~\ref{relation-to-kernel-density-estimation}.
Here, we provide theoretical estimates to illustrate the framework’s
scalability potential for global POI datasets.

\subsubsection{Time Complexity}
\label{time-complexity}

The standard brute-force approach for calculating influence across \(n\)
POIs would require \(O(n)\) distance computations per query point. In
contrast, the FAISS IVF index reduces this to: 
\begin{equation}
\label{time-complexity}
O(n_{probe} \cdot \frac{n}{n_{list}} + k)
\end{equation}
where \(n_{probe}\) is the number of centroids searched, \(n/n_{list}\)
represents the average size of an inverted list, and \(k\) is the cost
of sorting the final candidates. For large-scale datasets where
\(n_{list} \approx \sqrt{n}\), this effectively transforms the search
into sub-linear time, enabling real-time influence field construction
for high-frequency trajectory data.

\subsubsection{Memory Complexity}
\label{memory-complexity}

The memory overhead is dominated by the inverted lists storing the ECEF
coordinates. The complexity is:
\begin{equation}
\label{memory-complexity}
O(n \cdot d)
\end{equation}
where \(d=3\) for our 3D
spatial vectors. Given that each POI requires around 12 bytes (3 floats),
the framework is theoretically capable of accommodating millions of POIs
on commodity hardware with several gigabytes of RAM.

\emph{This demonstrates that the ADF framework balances expressive modeling
with computational practicality, making it suitable for global-scale
geospatial applications.}

\section{Case Study: Flight Trajectory POI Extraction}
\label{flight-trajectory-poi-extraction}

The ADF framework is applicable to a wide range of scenarios, which we
will discuss in the Section~\ref{discussion}. To illustrate the framework's
utility, we evaluate the extraction of points of interest (POIs) from flight
trajectories using the ADF method, compared against the supervised kinematic
baseline (the suggested grounded truth as labelled POI) introduced in the
Appendix~\ref{appendix}. In this setup a large flight trajectory dataset served as the reference
point set for POI extraction. A total of approximately 1.8 million POIs were
extracted from the dataset using the motion-based procedure described in
Section~\ref{baseline-poi-construction}. For evaluation, we employed a separate
dataset not included in the reference set. The extracted reference POIs were
then used to construct a local instance of the Adaptive Density Field (ADF) and
to illustrate POI extraction accross all the trajectories in the evaluation dataset.

\subsection{Baseline POI Construction}
\label{baseline-poi-construction}

To collect the POI data, we applied a physics-based motion residual
analysis pipeline as an example of 'manually labelled' POI baseline.
POIs are derived from trajectory segments where the motion model
fails to predict future positions accurately. To estimate future
aircraft positions, we applied a physics-based interpolation model
that blends two motion predictors:

\begin{enumerate}
\def\labelenumi{\arabic{enumi}.}
\item
  \textbf{Constant-Acceleration (CA) model:} reliable for nearly
  straight trajectories
\item
  \textbf{Cubic Hermite Spline interpolation:} smooth and accurate for
  curved motion
\end{enumerate}

the blending weight $w$ is an exponential decay function of the local
curvature $\kappa$ \cite{lee2007trajectory}, calibrated against
the 95th percentile of curvatures within each flight: Low curvature
indicates motion is nearly straight, hence CA weights more, and vice
versa. This adaptive combination produces a more stable and realistic
prediction than using either method alone.

To evaluate the quality of the predicted positions, we computed the
time‑normalized Mahalanobis loss \cite{mahalanobis1936distance} for
each flight. This metric captures not only the magnitude of prediction
errors but also their directional structure, covariance, and temporal
spacing. Then we labelled the Points of Interest (POIs)---locations where
the prediction error is unusually high \cite{chandola2009anomaly}. These
points often correspond to sharp maneuvers, abnormal motion, or sensor
irregularities, and they serve as valuable markers for downstream analysis.
Loss scores are normalized to extract POIs, and the threshold in this case
study is the \(75\%\) percentile.

However, in this case, the detected POI does not necessarily correspond to an
actual infrastructure feature; it simply marks a point where the motion
deviates significantly from our prediction. It should therefore be understood
as an example of task-specific POI definition tailored to aircraft motion.

\begin{quote}
\textbf{
Full implementation details and parameter settings are provided
in the \emph{Appendix~\ref{appendix}}.
}
\end{quote}

\subsection{ADF Framework and FAISS Acceleration}
\label{adf-framework-and-faiss-acceleration}

We instantiated the ADF framework on the nationwide POI dataset by
specifying the coordinate system, kernel parameterizations, and neighbor
search configurations. All POI locations were converted from WGS84
\cite{united1987department} geodetic coordinates to Earth-Centered,
Earth-Fixed (ECEF) Cartesian coordinates to ensure metric consistency.
POI scores were treated as non-negative scalar probability attention
values and were used directly without additional normalization.

For the scalable neighbor retrieval, we implemented a FAISS-based Inverted
File (IVF) index structure over the ECEF coordinates. The spatial domain
was partitioned into $4096$ (\(2^{12}\)) Voronoi cells (clusters) during
the training phase using a flat L2 quantizer. This partitioning scheme
optimizes the search space by narrowing the query range to specific geographic
regions rather than entire nationwide dataset. During the inference phase,
we configured the nprobe parameter to $16$, meaning that for each query
point, only the $16$ nearest Voronoi cells, those with centroids closest to
the query location were probed. This configuration allows for high-fidelity
approximate nearest neighbor (ANN) retrieval of the top $k=100$ neighbors
while maintaining the computational efficiency required for processing millions
of trajectory points.

Crucially, the global spatial scale parameter was fixed to $\sigma_0=500$
meters for all experiments, providing a consistent baseline for the kernel
bandwidth. The ADF values were then evaluated strictly using the formulation
defined in Section 2, ensuring that the adaptive kernel
$(\sigma=\sigma_0 / (scores + 1e-6))$ remained mathematically consistent
with the established framework without further modification.

\begin{quote}
Unless otherwise stated, these parameter choices were held constant
throughout the case study.
\end{quote}

\subsection{Evaluation and POI Extraction}
\label{evaluation-and-poi-extraction}

To analyze the ADF along individual trajectories in Chengdu region, we:

\begin{enumerate}
\def\labelenumi{\arabic{enumi}.}
\item
  \textbf{Convert the coordinates:} Transform all points from WGS84
  geodetic coordinated to ECEF Cartesian coordinates.
\item
  \textbf{Evaluate ADF:} Compute the \(ADF(x_t)\) at each trajectory
  point using the FAISS-accelerated framework with adaptive bandwidth
  defined in Section 2.
\item
  \textbf{Re-attach geodetic coordinates:} Join the computed ADF values
  back to the original $(lon, lat, alt, t)$ records for downstream
  spatial-temporal analysis.
\end{enumerate}

The extraction of final POIs from the density field follows the principle
that influence should be assessed \emph{relative to operational context}
rather than a rigid global threshold \cite{zheng2015trajectory}. Because
aircraft operate across varying density regimes (e.g., congested terminal
areas versus sparse cruise sectors), we employ a trajectory-specific relative
threshold.Let $\{F_t\}$ denote the sequence of ADF values along a single
trajectory. A trajectory point at time $t$ is labeled as a POI if:
$$F_t \ge P_{75}(\{F_t\})$$
where $P_{75}$ represents the 75th percentile of field intensities
experienced by that specific flight, this percentile was selected to maintain
consistency with the baseline prevalence discussed in
Section~\ref{baseline-poi-construction}. As demonstrated in
Section~\ref{ablation-study}, utilizing a relative threshold ensures
statistical parity between the baseline and ADF results, allowing for a
one-to-one comparison of the detection fidelity across the evaluation
dataset.

This adaptive criterion yields a \emph{binary mask along each trajectory},
effectively normalizing the detection process against the "background noise"
of the flight's environment. Consequently, a flight primarily traversing
low-density regions can still exhibit localized POIs (e.g., a sudden
maneuver in mid-air), while a flight in a persistently congested terminal
area must exceed its own higher-than-average baseline to trigger a detection.
This approach ensures high specificity and prevents the "hallucination"
of POIs in areas of high global density that lack local behavioral
significance.

\subsection{Comparative Validation: ADF vs. Kinematic Baseline}
\label{comparative-validation}

The comparative validation demonstrates a high degree of spatial consistency
between the ADF framework and the kinematic baseline. Notably, as the threshold
increases from 150 m to 300 m, the ADF's matched points rise from 16,657 to
18,074, representing a consistency rate of approximately $80\%$ relative to its
own total POI count ($20,769$). The observed ratio suggests that the ADF method
effectively resolves coarse kinematic anomalies into fine-grained behavioral
clusters.

As the ADF and kinematic baseline rely on distinct mathematical foundations,
a strict coordinate-wise overlap is not expected. Instead, the validation
assesses spatial concordance within a specified distance threshold. To ensure
a fair comparative analysis, we enforced cardinality matching between the two
sets; where the ADF generated multiple candidate points for a single kinematic
event, the redundant labels were consolidated to ensure the total population
(\(N\)) remained consistent across both methods. This normalization allows
us to test the model's labeling accuracy without the results being skewed
by differences in point density.

\begin{table*}[htbp]
\centering
\caption{Comparative Spatial Matching: Baseline vs. ADF (Fine-Scale Sensitivity)}
\label{tab:validation-sensitivity}
\begin{tabularx}{\linewidth}{@{} l XX XX XX @{}}
\toprule
& \multicolumn{2}{c}{\textbf{150 meters}} & \multicolumn{2}{c}{\textbf{200 meters}} & \multicolumn{2}{c}{\textbf{300 meters}} \\
\cmidrule(lr){2-3} \cmidrule(lr){4-5} \cmidrule(lr){6-7}
\textbf{Category} & Base & ADF & Base & ADF & Base & ADF \\
\midrule
\textbf{Matched (TT)} & \multicolumn{2}{c}{\textbf{16,606}} & \multicolumn{2}{c}{\textbf{17,225}} & \multicolumn{2}{c}{\textbf{17,953}} \\
\textbf{Unique (TF/FT)} & 10,758 & 4,163 & 10,492 & 3,544 & 10,056 & 2,816 \\
\midrule
\textbf{Precision (\%)} & \multicolumn{2}{c}{\textbf{79.96\%}} & \multicolumn{2}{c}{\textbf{82.94\%}} & \multicolumn{2}{c}{\textbf{86.44\%}} \\
\bottomrule
\end{tabularx}
\footnotesize{Note: Precision is defined as the ratio of matched points to total points ($N_{ADF}=20,769$).}
\end{table*}

\begin{figure}[htbp]
    \centering
    \includegraphics[width=\columnwidth]{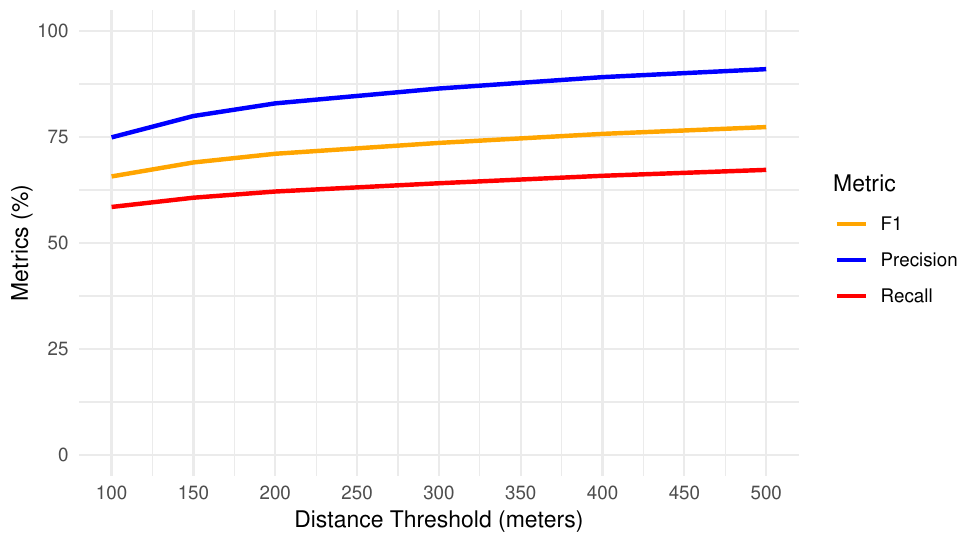}
    \caption{
      \textbf{ADF Spatial Fidelity Sensitivity Analysis.}
      This plot illustrates the performance metrics of the ADF
      framework, including Precision, Recall, and F1-score, relative
      to the kinematic baseline across an expanding distance threshold
      (100 m to 500 m). The asymptotic increase in precision beyond
      400 m demonstrates that the ADF method anchors to kinematic
      behavioral anomalies while maintaining high spatial concordance,
      even as the "soft" matching criteria are relaxed.
    }
    \label{fig:Fig.2}
\end{figure}

While Table \ref{tab:validation-sensitivity} details the precise point
counts for fine-scale thresholds, Figure \ref{fig:Fig.2}
illustrates the broader convergence of the two methods. As the spatial
tolerance is relaxed to 500 m, the ADF precision plateaus at above
\(90\%\), confirming that the vast majority of ADF-extracted POIs are
spatially anchored to kinematic anomalies, even if they exhibit higher
local precision than the baseline at the 100 m scale.

Performance improves smoothly with increasing spatial tolerance, with
precision exceeding $80\%$ at 200 m and stabilizing above $90\%$ beyond
400 m, while recall (defined as baseline-covered ADF detections) remains
consistently bounded around $60\sim70\%$, indicating stable spatial
localization rather than threshold-dependent artifacts.

\begin{figure}[htbp]
    \centering
    \includegraphics[width=\columnwidth]{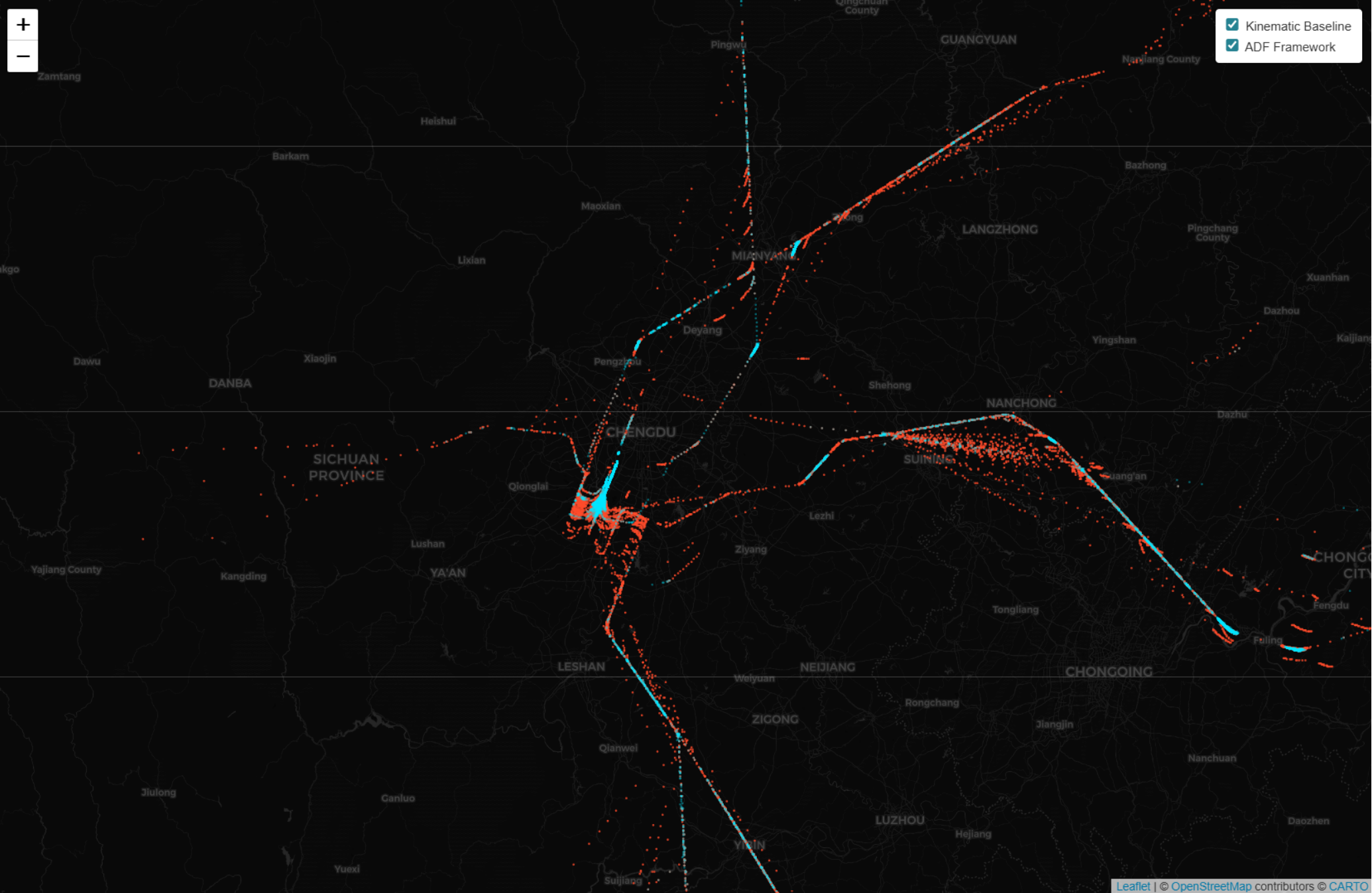}
    \caption{
      \textbf{Spatial Distribution of Extracted POIs in Chengdu Region.}
      This 2D projection illustrates the geographic distribution of
      points identified by the ADF framework (cyan) and the kinematic
      baseline (red). While all computations in
      Section~\ref{flight-trajectory-poi-extraction}, including neighbor
      retrieval and density field construction, were performed in 3D
      ECEF Cartesian space to ensure metric integrity, the results are
      visualized here in 2D geodetic coordinates for regional context.
      The significant overlap in high-traffic corridors and terminal
      areas near CTU visually confirms the spatial concordance
      quantified in Table~\ref{tab:validation-sensitivity}.
    }
    \label{fig:Fig.3}
\end{figure}

It is important to emphasize that while Figure~\ref{fig:Fig.3} provides
a 2D planimetric view of the results for visualization purposes, all
underlying experiments and proximity searches were executed in the full
3D ECEF coordinates system. This ensures that the vertical dimension of
flight trajectories is fully accounted for in the ADF mathematical model,
even when projected onto a 2D map.

\section{Ablation Study}
\label{ablation-study}

To evaluate the robustness and efficiency of the proposed framework, a
series of ablation experiments were conducted. This study systematically
isolates the key components of the system, computational acceleration via
indexing, kernel bandwidth strategies, and hyperparameter sensitivity to
quantify their contribution to the overall performance. By comparing the
full implementation against simplified or non-optimized baselines, we
establish the empirical justification for the architectural choices
detailed in Section~\ref{flight-trajectory-poi-extraction}.

\subsection{On Score and Bandwidths}
\label{on-score-and-bandwidths}

A critical feature of the ADF is its use of adaptive bandwidths ($\sigma$)
and POI scores ($s$). To verify the necessity of this complexity, the model
was compared against a standard fixed-bandwidth Kernel Density Estimation
(KDE) where $\sigma$ was held constant (150 and 200) and $s$ was ignored
(effectively treating all POIs as equally significant).

\begin{table}[htbp]
\centering
\caption{Comparison table: fixed and adaptive bandwidths (150m threshold)}
\label{tab:fixed-and-adaptive-bandwidths-1}
\newcolumntype{Y}{>{\Centering\arraybackslash}X}
\begin{tabularx}{\linewidth}{@{} >{\hsize=1\hsize}Y >{\hsize=1\hsize}Y >{\hsize=1\hsize}Y >{\hsize=1\hsize}Y @{}}
\toprule
\textbf{Bandwidth $\sigma$} & \textbf{Precision} & \textbf{Recall} & \textbf{F1} \\
\midrule
\textbf{adaptive} & $79.96\%$ & $60.69\%$ & $69.00\%$ \\
\textbf{250} & $76.98\%$ & $61.04\%$ & $68.09\%$ \\
\textbf{500} & $79.77\%$ & $60.57\%$ & $68.86\%$ \\
\textbf{750} & $79.84\%$ & $60.60\%$ & $68.91\%$ \\
\bottomrule
\end{tabularx}
\end{table}

In Table \ref{tab:fixed-and-adaptive-bandwidths-1} and
\ref{tab:fixed-and-adaptive-bandwidths-2}, we demonstrated the comparison
between fixed bandwidth (250, 500, 750 meters) and adaptive bandwidth under
150 and 200 meters tolerance threshold (The nprobe for FAISS is set to 16,
and k=100). While fixed bandwidths in some cases can achieve higher recall by
capturing smaller, isolated POI clusters, they all suffer from a slight
reduction in precision whose intensity is affected by the bandwidth settings,
which make the output unstable and hard to predict. The Adaptive ADF utilizes
local POI scores to modulate the kernel width, which provides a superior balance
between spatial specificity and sensitivity, resulting in the stability of
the precision and F1 score.

\begin{table}[htbp]
\centering
\caption{Comparison table: fixed and adaptive bandwidths (200m threshold)}
\label{tab:fixed-and-adaptive-bandwidths-2}
\newcolumntype{Y}{>{\Centering\arraybackslash}X}
\begin{tabularx}{\linewidth}{@{} >{\hsize=1\hsize}Y >{\hsize=1\hsize}Y >{\hsize=1\hsize}Y >{\hsize=1\hsize}Y @{}}
\toprule
\textbf{Bandwidth $\sigma$} & \textbf{Precision} & \textbf{Recall} & \textbf{F1} \\
\midrule
\textbf{adaptive} & $82.94\%$ & $62.15\%$ & $71.05\%$ \\
\textbf{250} & $79.55\%$ & $62.44\%$ & $69.96\%$ \\
\textbf{500} & $82.63\%$ & $61.97\%$ & $70.83\%$ \\
\textbf{750} & $82.84\%$ & $62.06\%$ & $70.96\%$ \\
\bottomrule
\end{tabularx}
\end{table}

\subsection{On Implementation of FAISS}
\label{on-implementation-of-faiss}

To validate the efficiency of the proposed system, a comparative analysis
was conducted between the optimized FAISS-based implementation and a
traditional brute force approach. This experiment aimed to quantify
the computational gains while ensuring that the spatial partitioning used
by FAISS did not degrade the integrity of the proposed ADFs.

\begin{table}[htbp]
\centering
\caption{Comparison table: ADF with FAISS vs. without FAISS}
\label{tab:faiss-or-no-faiss}
\begin{tabularx}{\linewidth}{@{} l cccc @{}}
\toprule
\text{FAISS} & \textbf{Precision} & \textbf{Recall} & \textbf{F1} & \textbf{ms/query} \\
\midrule
\textbf{yes} & $82.94\%$ & $62.15\%$ & $71.05\%$ & $0.1145$ \\
\textbf{no} & $82.94\%$ & $62.15\%$ & $71.05\%$ & $10.8184$ \\
\bottomrule
\end{tabularx}
\end{table}

As summarized in Table \ref{tab:faiss-or-no-faiss}, the implementation of
FAISS (using an IVF-Flat index with \textit{nprobe}=16) yielded results
identical to the Brute Force method (under tolerance threshold of 200 meters)
across all primary performance metrics: Precision (82.94\%), Recall (62.15\%),
and F1-score (71.05\%). This demonstrates that the approximation error of
the IVF index is negligible under a threshold of 200 m in this case,
successfully capturing the relevant neighbor set ($k=100$) required for
accurate density estimation.

The most significant finding lies in the computational throughput. The Brute
Force approach required approximately 10.82 ms per query point, whereas the
FAISS-optimized system reduced this to 0.11 ms. This represents a nearly
100-fold increase in processing speed. Such a reduction is critical for real-time
aviation applications, transforming a process that would take minutes into one
that executes in seconds, thereby enabling the scalable analysis of large-scale
flight datasets.

\subsection{On nprobe and k}
\label{on-nprobe-and-k}

\begin{table}[htbp]
\centering
\caption{Comparison table: nprobe selection}
\label{tab:nprobe-selection}
\newcolumntype{Y}{>{\Centering\arraybackslash}X}
\begin{tabularx}{\linewidth}{@{} l cccc @{}}
\toprule
\textbf{probe} & \textbf{Precision} & \textbf{Recall} & \textbf{F1} & \textbf{ms/query} \\
\midrule
\textbf{4} & $82.92\%$ & $62.14\%$ & $71.04\%$ & $0.1255$ \\
\textbf{8} & $82.94\%$ & $62.15\%$ & $71.05\%$ & $0.1484$ \\
\textbf{16} & $82.94\%$ & $62.15\%$ & $71.05\%$ & $0.1596$ \\
\textbf{64} & $82.94\%$ & $62.15\%$ & $71.05\%$ & $0.2920$ \\
\textbf{256} & $82.94\%$ & $62.15\%$ & $71.05\%$ & $0.9571$ \\
\bottomrule
\end{tabularx}
\footnotesize{Note: The threshold is set to 200 meters, k=100.}
\end{table}

In this section, we evaluate the impact of the index search depth, defined
by the \textit{nprobe} parameter, on both the classification performance
and computational efficiency of the Adaptive Density Function (ADF). The
parameter $k$ was fixed at 100 to ensure a sufficiently large neighborhood
for the local density kernels, while \textit{nprobe} was varied from 4 to
256 to observe the approximation error introduced by the Inverted File (IVF)
structure.

As demonstrated in Table \ref{tab:nprobe-selection}, the
classification metrics (Precision, Recall, and F1-score) remain remarkably
stable across different \textit{nprobe} values. Even at a low search depth
of \textit{nprobe}=8, the system achieves an F1-score of 71.05\%, identical
to the score at the highest search depth of 256. This suggests that the
high-density regions identifying Points of Interest (POIs) are sufficiently
distinct that even a coarse approximate search effectively captures the
primary density contributors.

However, the computational cost scales significantly with the search depth.
While \textit{nprobe}=4 provides a latency of approximately 0.1255 ms per
query, increasing the depth to 256 raises the latency to 0.9571 ms, a nearly
eight-fold increase in processing time for negligible gains in classification
accuracy. It is worth noting that at very low \textit{nprobe} values,
diminishing returns in speed were observed, likely due to fixed computational
overheads in the Python-C++ interface. Based on these results,
\textit{nprobe}=16 was selected as the optimal operating point for the
remainder of the study, providing the best compromise between mathematical
fidelity and real-time processing requirements.

\begin{table}[htbp]
\centering
\caption{Comparison table: k value selection}
\label{tab:k-selection}
\newcolumntype{Y}{>{\Centering\arraybackslash}X}
\begin{tabularx}{\linewidth}{@{} l cccc @{}}
\toprule
\textbf{k} & \textbf{Precision} & \textbf{Recall} & \textbf{F1} & \textbf{ms/query} \\
\midrule
\textbf{50} & $82.97\%$ & $62.10\%$ & $71.04\%$ & $0.1435$ \\
\textbf{100} & $82.94\%$ & $62.15\%$ & $71.05\%$ & $0.1596$ \\
\textbf{150} & $82.95\%$ & $62.16\%$ & $71.07\%$ & $0.1724$ \\
\bottomrule
\end{tabularx}
\footnotesize{Note: The threshold is set to 200 meters, nprobe=16.}
\end{table}

Following the index optimization, the neighborhood size $k$ was evaluated
to determine its influence on the ADF smoothing effect and classification
accuracy. As shown in Table \ref{tab:k-selection}, the $k$ parameter was
varied from 50 to 150 with \textit{nprobe} fixed at 16. Similar to the
search depth findings, the system displays high resilience to changes in
$k$, with the F1-score fluctuating by less than 0.03\%. This stability
indicates that the regional density peaks are sufficiently dense that the
core POI classification is not highly sensitive to the peripheral
neighbors included in the kernel summation.

A minor increase in precision and F1-score was observed when moving from
$k=50$ to $k=150$, while precision remained remarkably stable (fluctuating
by less than 0.03\%). suggesting that a larger neighborhood helps bridge
small gaps in the density field, resulting in more cohesive POI
extraction. However, increasing $k$ beyond 100 yielded diminishing
returns, with a linear increase in computational latency from 0.14 ms
to 0.17 ms per query. This latency increase is attributed to the higher
dimensionality of the distance and score vectors processed during the
kernel computation phase. Consequently, $k=100$ was finalized as the
optimal neighborhood size based on the speed and F1-score as it
maximizes recall without incurring the over-smoothing or computational
costs associated with higher $k$ values.

Ultimately, the combination of \textit{nprobe}=16 and $k=100$ establishes
a high-performance configuration capable of processing trajectory points
at sub-millisecond speeds while maintaining the mathematical fidelity of
a brute-force approach.

\section{Comparative Evaluation and Related Works}
\label{comparative-evaluation-and-related-works}

\subsection{A Simple Comparison with Classical k-Nearest-Neighbors method(KNN)}
\label{a-simple-comparison-with classical-knn-method}

For the benchmark comparison study with existing traditional methods, we compared
the proposed framework with the standard K-Nearest-Neighbors (KNN) as a canonical
baseline.

The K-Nearest-Neighbors (KNN) baseline was implemented as an unsupervised density
estimator operating in three-dimensional Earth-Centered, Earth-Fixed (ECEF)
coordinates. This method serves as a purely geometric proximity baseline, where
the local density at any trajectory point is inversely proportional to the mean
distance of its  closest neighbors in the reference POI dataset. Unlike supervised
classification, this approach identifies Points of Interest (POIs) by selecting
trajectory points that exhibit the highest local density (lowest mean distance)
relative to the global distribution of the flight.

As implemented in the experimental pipeline, flight coordinates are first
transformed to Cartesian ECEF space to ensure consistent Euclidean distance
metrics across varying flight levels. For each query point, the algorithm
identifies the set of neighbors and computes the density proxy as:
\[
D(p) = \frac{1}{k} \sum_{q \in \mathcal{N}_k(p)} |p - q|_2
\]
where lower values of \(D(p)\) correspond to higher local density. A binary POI
mask is then generated by applying a threshold at the 75th percentile of the
observed distances (selecting the 25\% of points with the highest proximity). The
performance of this baseline was evaluated across a sweep of  values to assess
sensitivity to the neighborhood scale.

\begin{table}[htbp]
\centering
\caption{KNN Baseline Performance Metrics (3D ECEF, 200m Threshold)}
\label{tab:knn-bench}
\begin{tabularx}{\linewidth}{@{} l cccc @{}}
\toprule
\textbf{k} & \textbf{Precision} & \textbf{Recall} & \textbf{F1-Score} & \textbf{ms/query} \\
\midrule
25 & 53.91\% & 95.40\% & 68.89\% & 0.3671 \\
50 & 53.70\% & 95.09\% & 68.64\% & 0.3361 \\
75 & 53.46\% & 94.81\% & 68.37\% & 0.3499 \\
500 & 51.31\% & 92.22\% & 65.93\% & 0.4753 \\
\bottomrule
\end{tabularx}
\end{table}

The quantitative results in Table \ref{tab:knn-bench} highlight a significant
trade-off in the KNN approach. While the algorithm achieves an exceptionally high
recall (exceeding 95\% at $k=25$), it suffers from poor precision
($\approx50\sim55\%$). This behavior results in a large number of False Positives
(exceeding 28,000 points per run), suggesting that while geometric proximity is a
necessary condition for identifying POIs, it is not a sufficient one; the lack of
score-modulated attention in KNN leads to the inclusion of high-density but
low-significance background points. Furthermore, the increase in k to 500 leads
to a degradation in both precision and recall, confirming that over-extending the
neighborhood window can introduce excessive noise into the local density estimation.

\subsection{Relation to Kernel Density Estimation}
\label{relation-to-kernel-density-estimation}

Classical KDE \cite{silverman1986density} evaluates a global
sum over all points using fixed or locally adaptive bandwidths.
In contrast, the Adaptive Density Field (ADF) is a query‑conditioned,
local aggregation operator: for each query location ADF selects a
bounded neighbor set (via k‑NN retrieval) and aggregates score‑modulated
kernels centered on those neighbors.

Although ADF shares mathematical components with adaptive KDE
\cite{terrell1992variable} (both use variable bandwidths and local weighting),
their objectives differ: adaptive KDE methods estimate probability densities
from samples and typically rely on all observations, whereas ADF constructs an
application‑driven influence field. In ADF, bandwidths are modulated by externally
supplied influence scores (or learned mappings) and the kernel support is defined
by the query‑dependent neighbor set, so the operator preserves spatial structure
and application semantics rather than producing a normalized probability density.

Consequently, ADF does not estimate a traditional probability density, but
constructs a continuous influence field that preserves spatial structure and
application-specific attention scores. The table shown below summarizes
the direct comparison between variable‑bandwidth KDE and the proposed ADF.


ANN-accelerated KDE method \cite{swand1994fast} primarily focus on speeding
up density estimation. In contrast, ADF integrates approximation directly
into the operator: k‑NN selection serves as top‑k attention pruning rather
than a mere post hoc speedup. In this way, ADF reframes adaptive kernels
as a geometry‑preserving attention operator, where sparsification is
intrinsic and kernel choices and score-bandwidth mappings remain
design degrees of freedom.

\subsection{Relation to the Attention Mechanism}
\label{relation-to-the-attention-mechanism}

The proposed ADF method was inspired by the Attention Mechanism
\cite{vaswani2017attention}. While attention mechanisms are typically
defined over discrete tokens in a learned latent space, ADF operates
directly in continuous metric space. Similarity is induced by learned
physical distance rather than learned projections, and attention weights
arise from energy-based kernels instead of normalized dot products. As a
result, ADF should be viewed not as a rebranding of Transformer attention,
but as a geometric generalization of attention to continuous spatial domains.


This correspondence is structural rather than metaphorical:
\emph{ADF satisfies the abstract definition of attention as a
metric-induced aggregation mechanism over continuous space.}
The attention analogy is used as a structural lens rather than a claim
of architectural novelty.

\subsection{Other Related Works}
\label{other-related-works}

Pattern matching and spatial query systems return discrete matches under
complex constraints and rely on different indexing strategies (e.g., IR‑trees)
\cite{fang2019evaluating}, distinguishing them from continuous influence operators
like ADF. Control‑based adaptive retrieval (e.g., PIDKNN) adapts search radius
per query and offers an alternative to ANN retrieval \cite{qiao2022pid}.
Density‑based clustering methods such as OPTICS reveal hierarchical density
structure and avoid a single global density threshold, but they operate in
batch and output cluster labels rather than continuous, query‑conditioned
influence fields \cite{ester1997spatial}.

Adaptive KDE and k‑NN density estimators form the statistical backbone
for local bandwidth selection and smoothing \cite{scott2015multivariate}.
Hybrid kNN–kernel methods have been applied to spatio‑temporal clustering and
activity detection \cite{musdholifah2010knn, koylu2019deep}, but they
operate as dataset‑level estimators rather than per‑query operators.
Some query‑conditioned influence algorithms and scalable spatial query systems
address per‑query efficiency \cite{qi2018continuous, alghushairy2020genetic}, yet
they typically define influence via density contribution or discrete matching
rather than as a score‑modulated attention operator. ANN‑accelerated KDE pipelines
demonstrate practical scalability on large datasets \cite{thompson2022ancient}, but
in those works ANN is an implementation detail rather than an intrinsic component
of the operator. ADF differs by (i) treating neighbor selection and sparsification
as part of the operator, (ii) modulating kernel bandwidths with external scores,
and (iii) framing the aggregation as geometry‑preserving attention.
By embedding ANN-based sparsification into the operator itself, ADF addresses
the computational bottleneck of high-resolution density modeling in a way that
remains theoretically consistent with geographic principles of locality.

\subsection{Why This Matters}
\label{why-this-matters}

The suggested ADF is best understood as a \emph{geometric attention mechanism:}
a query‑conditioned, metric‑driven aggregation over continuous space.
In this sense, ADF implements a \emph{spatial attention operator} rather than
a traditional kernel estimator. While ADF shares mathematical components with
adaptive KDE and ANN-accelerated density estimation, its contribution is not a
new kernel estimator per se. Instead, ADF explicitly formulates these components
as a geometric attention operator, where approximation, sparsification, and
metric structure are intrinsic to the definition rather than implementation
details. Overall, ADF should be understood as a framework-level operator
rather than a fixed algorithm, with emphasis on structure, scalability, and
geometric grounding.

\section{Discussion}
\label{discussion}

The proposed Adaptive Density Field (ADF) framework provides a flexible,
scalable mechanism for aggregating sparse, heterogeneous points of
interest (POIs) into a continuous spatial field. By coupling
trajectory-conditioned analysis with POI extraction as a downstream
instantiation of ADF, this methodology captures both globally recurrent
patterns and locally significant deviations.

\subsection{Future application and extensions}
\label{future-application-and-extensions}

\begin{enumerate}
\def\labelenumi{\arabic{enumi}.}
\item
  \textbf{Operational Airspace Management:} The ADF-POI pipeline we
  presented in the Section~\ref{flight-trajectory-poi-extraction} could
  assist air traffic controllers in identifying target regions depending
  on the specific POI definition, such as regions of frequent maneuvering
  or congestion, enabling data-driven decisions for route planning, holding
  pattern optimization, and and safety monitoring.
\item
  \textbf{Predictive Trajectory Analysis:} Beyond post-hoc evaluation,
  the framework can be integrated with predictive models for trajectory
  planning, anomaly detection, or risk exposure assessment. By defining
  the POIs, it is possible to derive trajectory-conditioned POIs to
  provide adaptive thresholds for identifying unusual events relative to
  expected motion patterns.
\item
  \textbf{Cross-Domain Applicability:} While this study focuses on
  aircraft trajectories due to the initial problem setting, the
  methodology is domain-agnostic. Any scenario involving discrete
  spatiotemporal observations with underlying metric regularities,
  as long as the spatial characteristics are stable and fixed
  (in other words, not dynamic), such as maritime traffic, pedestrian
  movement, vehicle flows, or wildlife tracking -- we expect ADF-based
  spatial attention can identify zones of concentrated
  activities.
\item
  \textbf{Integration with Semantic Context:} Future work could embed
  semantic labels, such as airspace type, weather conditions, or
  operational procedures, into the ADF formulation, yielding
  geo-semantic density fields. This way, interestingly, aligns even more
  closely with the application of the attention mechanism in
  spatiotemporal analysis.
\item
  \textbf{Real-Time Applications:} With optimizations to indexing,
  neighbor search, or GPU-accelerated computation, the ADF framework
  could support near-real-time monitoring of live trajectories in
  drones, robotics, and other autonomous systems, allowing dynamic POI
  identification and interactive visualization for operational
  decision-making.
\end{enumerate}

\subsection{Methodological Reflections}
\label{methodological-reflections}

\begin{itemize}
\item
  The trajectory-conditioned approach ensures that high-density areas
  are interpreted relative to the agent's path, avoiding misleading
  conclusions based solely on absolute field intensity.
\item
  Parameter choices, such as kernel bandwidth and dominance factor,
  influence sensitivity and specificity; future studies could explore
  adaptive or data-driven tuning to optimize POI detection across
  heterogeneous datasets.
\item
  Visualization remains crucial for interpretation: layered maps
  combining POIs, ADF intensity, and POI labels provide immediate
  insight into both recurrent behavior and outlier events, but
  higher-dimensional visualizations or interactive dashboards could
  further enhance understanding.
\item
  This study represents a conceptual framework, and some details may
  therefore be approximate, incomplete, or provisional.
\end{itemize}

\section{Conclusion}
\label{conclusion}

This work introduces Adaptive Density Fields (ADF) as a
formulation-level geometric attention operator for scalable aggregation
of sparse spatial events in continuous metric space. By defining spatial
influence through query-conditioned neighbor selection, score-modulated
kernel weighting, and approximate nearest-neighbor execution, ADF
reframes adaptive kernel aggregation as an intrinsic form of
geometry-preserving attention rather than a purely statistical
estimator. The primary contribution lies not in a specific kernel choice
or application, but in the structural decomposition of spatial
aggregation: locality induced by k-NN attention, metric-grounded
weighting, and approximation treated as a defining component of the
operator. This perspective unifies ideas from adaptive kernel methods,
spatial indexing, and attention mechanisms under a common geometric
framework that scales to millions of points while remaining
interpretable.

To demonstrate the utility of this formulation, we presented an example
instantiation in the context of aircraft trajectory analysis.
Motion-derived Points of Interest were aggregated into a continuous ADF,
and trajectory-conditioned Points of Interest (POIs) were extracted
using a relative dominance criterion along individual paths. The case
study illustrates how the resulting fields reveal recurrent airspace
structures while distinguishing localized, trajectory-specific
deviations. Beyond aviation, the ADF framework is domain-agnostic and
applicable to a wide range of spatial and spatiotemporal settings,
including urban mobility, maritime traffic, robotics, and environmental
monitoring. Future work will explore semantic augmentation of the field,
adaptive parameter learning, anisotropic kernels, and real-time
deployment, further extending the applicability of geometric attention
in continuous spatial systems. Overall, ADF provides a robust and
extensible foundation for scalable, interpretable spatial attention,
bridging geometric structure and efficient computation in complex
spatial environments.

\section*{Acknowledgement}

This manuscript benefited from generative AI tools in limited,
non-substantive ways. ChatGPT (version 4o-mini) \cite{openai2024chatgpt}
and Gemini 3 \cite{google2026gemini} were used to improve language clarity
and fluency. Microsoft Copilot (search mode) \cite{microsoft2026copilot}
was employed to assist with citation discovery and verification. All
conceptual content, analysis, and argumentation were developed by the
author.

\bibliographystyle{plain}
\bibliography{references}
\clearpage

\appendix

\section{Physics-informed Trajectory POI Detection
Pipeline}
\label{appendix}

\textit{The appendix provides implementation-level details and mathematical
formulations supporting the methods described in the main text. Citations
follow the same reference list as the main body.}

\subsection*{1. Preprocessing the Flight
Data}
\label{preprocessing-the-flight-data}

\subsubsection*{1.1. Coordinate Conversion \cite{hofmann2012global}: WGS84 Geodetic
to ECEF}
\label{coordinate-conversion-wgs84-geodetic-to-ecef}

Given the latitude \(\varphi\) (rad), longitude \(\lambda\) (rad),
ellipsoidal height \(h\) (m) and the WGS84 parameters:

First compute the prime vertical radius of curvature:

\[
\tag{1.1.1}
N(\varphi) = \frac{a}{\sqrt{1 - e^2 \sin^2\varphi}}
\]

Then ECEF coordinates \((x,y,z)\):

\[
\tag{1.1.2}
\begin{aligned}
x &= \left(N(\varphi) + h\right)\cos\varphi\cos\lambda \\
y &= \left(N(\varphi) + h\right)\cos\varphi\sin\lambda \\
z &= \left(N(\varphi)(1 - e^2) + h\right)\sin\varphi
\end{aligned}
\]

Hence we get the ENU coordinates.

\subsubsection*{1.2. Coordinate Conversion: ECEF to ENU Conversion}
\label{coordinate-conversion-ecef-to-enu-conversion}

Pick a reference point (the origin of the local ENU frame in this case)
with geodetic coordinates \((\varphi_0,\lambda_0,h_0)\), and compute its
ECEF coordinates \((x_0,y_0,z_0)\) using the same equations as above.

For any point with ECEF \((x,y,z)\), define the difference vector:

\[
\tag{1.2.1}
\begin{bmatrix}
\Delta x \\ \Delta y \\ \Delta z
\end{bmatrix}=\begin{bmatrix}
x - x_0 \\ y - y_0 \\ z - z_0
\end{bmatrix}
\]

And given the Rotation matrix and ENU coordinate at reference
\((\varphi_0,\lambda_0,h_0)\):

\[
\tag{1.2.2}
\mathbf{R}=\begin{bmatrix}
\sin\varphi_0 & \cos\varphi_0 & 0 \\
\cos\varphi_0\cdot\sin\lambda_0 & -\sin\varphi_0\cdot\sin\lambda_0 & \cos\lambda_0 \\
\cos\varphi_0\cdot\cos\lambda_0 & \sin\varphi_0\cdot\cos\lambda_0 & \sin\lambda_0
\end{bmatrix}
\]

Therefore we have the calculation:

\[
\tag{1.2.3}
\begin{bmatrix}
E \\ N \\ U
\end{bmatrix}=\begin{bmatrix}
\Delta x \\ \Delta y \\ \Delta z
\end{bmatrix}\cdot\mathbf{R}
\]

This is the standard ECEF → ENU transformation used in geodesy and
navigation.

\subsubsection*{1.3. Creating a Dictionary}
\label{creating-a-dictionary}

To organize per‑flight data extracted from each GeoJSON file, we build a
dictionary where each flight ID maps to three lists:

\begin{itemize}
\tightlist
\item
  coords --- longitude, latitude, altitude
\item
  vel --- velocity components
\item
  dt --- timestamps
\end{itemize}

In practice, we use a defaultdict so each new flight\_id automatically
initializes this structure.

\subsection*{2. Position Prediction}
\label{position-prediction}

To estimate future aircraft positions, I applied a \textbf{physics‑based
interpolation model} that blends two motion predictors:

\begin{enumerate}
\def\labelenumi{\arabic{enumi}.}
\tightlist
\item
  \textbf{Constant‑Acceleration (CA) model} \cite{bar2001estimation}
   --- reliable for nearly straight trajectories\\
\item
  \textbf{Cubic Hermite Spline interpolation} \cite{de1978practical}
  --- smooth and accurate for curved motion
\end{enumerate}

The spline utilizes local velocity vectors as tangents at each waypoint,
providing a geometrically consistent path \cite{virtanen2020scipy} that
complements the CA model's acceleration-based predictions."

The blending weight is determined by the \textbf{local curvature} of the
trajectory: - Low curvature → motion is nearly straight → CA dominates\\
- High curvature → motion bends → spline dominates

This adaptive combination produces a more stable and realistic
prediction than using either method alone.

\subsubsection*{2.1. General Prediction}
\label{general-prediction}

For each flight:

\begin{itemize}
\tightlist
\item
  Convert raw coordinates into a consistent Cartesian frame\\
\item
  Compute velocity and approximate acceleration\\
\item
  Estimate local curvature \(k\) using
\end{itemize}

\[
\tag{2.1.1}
k = \frac{\lVert \mathbf{v} \times \mathbf{a} \rVert}{\lVert \mathbf{v} \rVert^3}
\]

\begin{itemize}
\tightlist
\item
  Compute a flight‑specific smoothing parameter
\end{itemize}

\[
\tag{2.1.2}
\alpha = \frac{\ln 5}{k_{95}}
\]

where \(k_{95}\) is the 95th percentile curvature\\
- For each timestamp, compute: - \textbf{Spline prediction} using
\texttt{CubicHermiteSpline} - \textbf{Constant‑acceleration prediction}
- Blend them using \(w = e^{-\alpha k}\):

\[
\tag{2.1.3}
\hat{p} = w \, p_{\text{CA}} + (1 - w) \, p_{\text{spline}}
\]

This yields a smooth, curvature‑aware prediction for each flight.

\subsubsection*{2.2. Loss Computation}
\label{loss-computation}

To evaluate the quality of the predicted positions, I compute a
\textbf{time‑normalized Mahalanobis loss} \cite{mahalanobis1936distance}
for each flight. This metric captures not only the magnitude of prediction
errors but also their \textbf{directional structure}, \textbf{covariance},
and \textbf{temporal spacing}.

The loss is computed in four main steps:

\paragraph{Extract Prediction Residuals}

For each flight, I compare the predicted positions \(\hat{p}_i\) with
the actual converted coordinates \(p_i\):

\[
\tag{2.2.1}
r_i = \hat{p}_i - p_i
\]

Only interior points are used \texttt{{[}2\ :\ size-2{]}} to avoid
boundary artifacts from the spline and acceleration models.

The residuals are then centered:

\[
\tag{2.2.2}
\tilde{r}_i = r_i - \bar{r}
\]

This removes global bias and ensures the covariance reflects
\emph{shape} rather than offset.

\paragraph{Estimate Residual Covariance}

The covariance of the centered residuals is computed as:

\[
\tag{2.2.3}
\Sigma = \operatorname{Cov}(\tilde{r}) + \lambda I
\]

A small Tikhonov regularization \cite{tikhonov1963solution}
term \(\lambda = 10^{-5}\) stabilizes the inversion of \(\Sigma\),
especially for nearly collinear trajectories.

The inverse covariance \(\Sigma^{-1}\) defines the \textbf{Mahalanobis
geometry} of the error space.

\paragraph{Compute Mahalanobis Distance}

For each residual vector:

\[
\tag{2.2.4}
d_i = \sqrt{\tilde{r}_i^\top \Sigma^{-1} \tilde{r}_i}
\]

This distance penalizes errors more strongly along directions where the
model is normally precise, and less along directions with naturally
higher variance.

\paragraph{Normalize by Temporal Spacing}

Because timestamps are not uniformly spaced, each error is scaled by a
time‑dependent factor:

\[
\tag{2.2.5}
t_i = \sqrt{\frac{\Delta t_i}{\bar{\Delta t}}}
\]

The final \textbf{time-relative Mahalanobis loss} is:

\[
\tag{2.2.6}
L_i = \frac{d_i}{t_i}
\]

This ensures that predictions made over longer time intervals are not
unfairly penalized compared to short‑interval predictions.

\subsection*{3. POI Detection}
\label{poi-detection}

After computing the time‑normalized Mahalanobis loss for each flight,
the next step is to identify \textbf{Points of Interest
(POIs)}---locations where the prediction error is unusually high. These
points often correspond to sharp maneuvers, abnormal motion, or sensor
irregularities, and they serve as valuable markers for downstream
analysis.

\textbf{However, a POI does not always represent an actual
infrastructure feature; it simply marks a point where the motion
deviates significantly.}

The POI detection pipeline consists of three main stages:

\subsubsection*{3.1. Normalize the Loss Scores}
\label{normalize-the-loss-scores}

For each flight, the Mahalanobis losses are rescaled to the interval
\texttt{[0,\ 1]}:

\[
\tag{3.1}
s_i = \frac{L_i - \min(L)}{\max(L) - \min(L) + \varepsilon}
\]

This normalization ensures that POI detection is \textbf{relative to
each flight's own dynamics}, making the method robust to differences in
scale, speed, or noise across flights.

\subsubsection*{3.2. Thresholding}
\label{thresholding}

Here, I introduced an element called POI score, which indicates how
anomalous each point is relative to the rest of the flight.

A point is flagged as a POI if its normalized score exceeds a fixed
threshold:

\[
\tag{3.2}
s_i \ge 0.75
\]

This threshold captures the upper quartile of anomalous behavior while
avoiding excessive false positives.

It can be adjusted depending on the desired sensitivity of the detection
process.

\subsubsection*{3.3. Export POIs to CSV}
\label{export-pois-to-csv}

Each detected POI is stored with: (i) flight ID; (ii) point index;
(iii) longitude, latitude, altitude; (iv) POI score

All POIs are aggregated into a Pandas DataFrame and exported as a CSV
file, enabling further visualization, inspection, or integration into
downstream workflows.

\end{document}